\documentclass[letterpaper, 10 pt, conference]{ieeeconf}  

\IEEEoverridecommandlockouts                              

\usepackage{epsfig} 

\usepackage[T1]{fontenc}
\usepackage{multirow}
\usepackage{cite}
\usepackage{flushend}
\usepackage{float}
\usepackage{CJKutf8}

\usepackage{subfigure}                 
\usepackage{threeparttable}
\usepackage[colorlinks=true, urlcolor=blue, linkcolor=red]{hyperref}

\title{\LARGE \bf
Triple-S: A Collaborative Multi-LLM Framework for Solving Long-Horizon Implicative Tasks in Robotics
}

\author{Zixi Jia\textsuperscript{1,*}, Hongbin Gao\textsuperscript{1}, Fashe Li\textsuperscript{2}, Jiqiang Liu\textsuperscript{1}, Hexiao Li\textsuperscript{1} and Qinghua Liu\textsuperscript{1}
\thanks{$^*$Corresponding author}%
\thanks{$^1$Zixi Jia, Hongbin Gao, Jiqiang Liu, Hexiao Li and Qinghua Liu are with the Faculty of Robot Science and Engineering, 
Northeastern University, Shenyang, 110819, China (email: jiazixi@mail.neu.edu.cn; 2302138@stu.neu.edu.cn; 2302152@stu.neu.edu.cn; 2410842@stu.neu.edu.cn; 2302155@stu.neu.edu.cn)}%
\thanks{$^2$Fashe Li is with SIASUN Robot Automation CO. Ltd., Shenyang, 110819, China (email: lifashe.in@siasun.com)}%
}

\begin{document}
\begin{CJK}{UTF8}{gbsn}

\maketitle
\thispagestyle{empty}
\pagestyle{empty}

\begin{abstract}
Leveraging Large Language Models (LLMs) to write policy code for controlling robots has gained significant attention. However, in long-horizon implicative tasks, this approach often results in API parameter, comments and sequencing errors, leading to task failure. To address this problem, we propose a collaborative Triple-S framework that involves multiple LLMs. Through In-Context Learning, different LLMs assume specific roles in a closed-loop Simplification-Solution-Summary process, effectively improving success rates and robustness in long-horizon implicative tasks. Additionally, a novel demonstration library update mechanism which learned from success allows it to generalize to previously failed tasks. We validate the framework in the Long-horizon Desktop Implicative Placement (LDIP) dataset across various baseline models, where Triple-S successfully executes 89$\%$ of tasks in both observable and partially observable scenarios. Experiments in both simulation and real-world robot settings further validated the effectiveness of Triple-S. Our code and dataset is available at: 
\url{https://github.com/Ghbbbbb/Triple-S}.
\end{abstract}

\section{INTRODUCTION}
Long-horizon tasks have been extensively explored in the robotics community, such as household tasks~\cite{c1,c2}, vision-language navigation~\cite{c3,c4}, and desktop object rearrangement~\cite{c5,c6}. Compared to standard long-horizon tasks, long-horizon implicative tasks are significantly more complex,  as they require not only proper sequencing of multiple actions (e.g., \textit{<move>} → \textit{<pick>} → \textit{<place>}) but also implicit reasoning over underspecified instructions (e.g., \textit{banana colored block} → \textit{yellow\_block}). Designing agents capable of addressing these challenges can fundamentally change the way robots integrate into our future society.

\begin{figure}[htp]
\begin{center}
\includegraphics[scale=0.32]{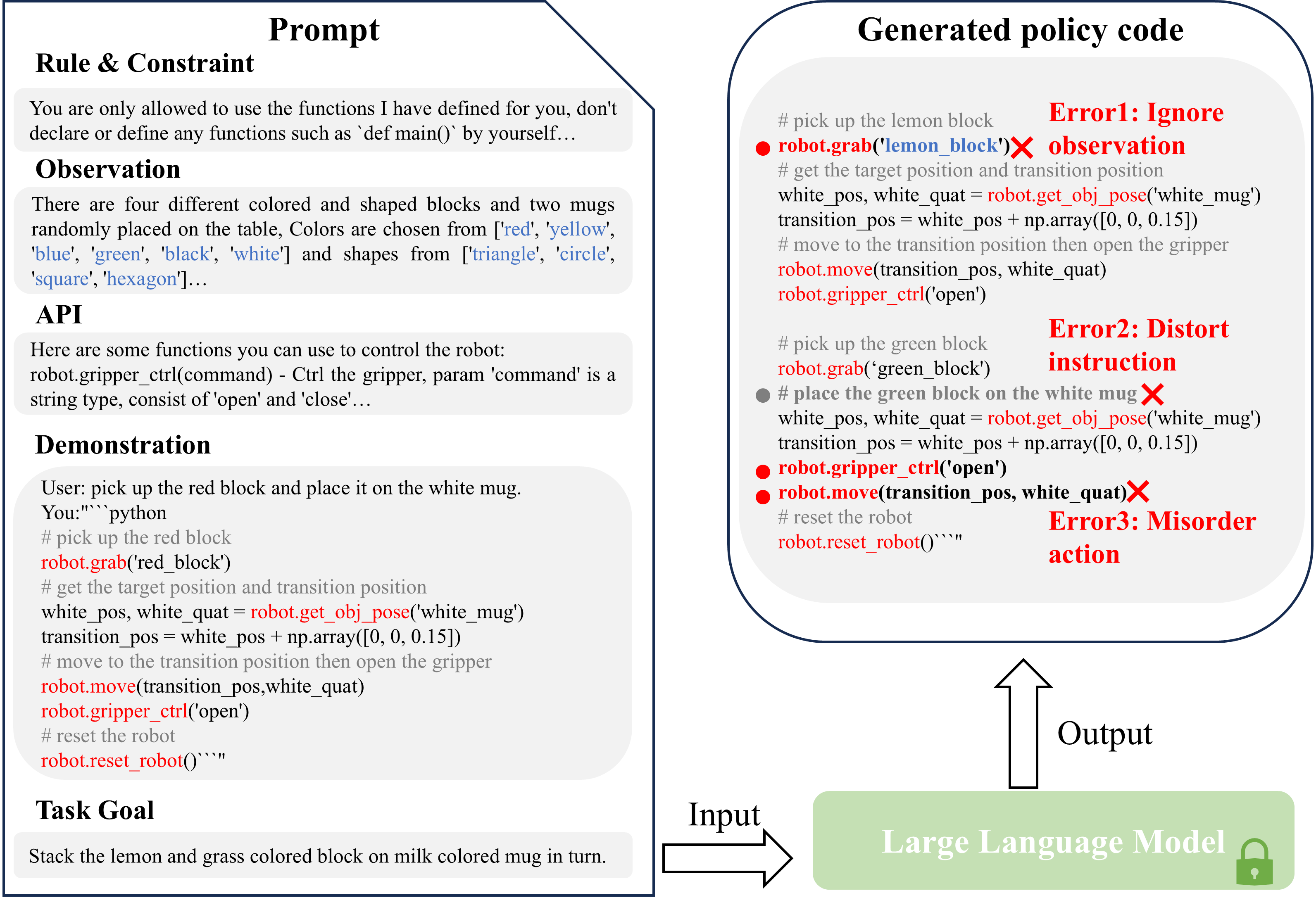}
\caption{Issues with the current policy code approach when predicting long-horizon implicative tasks (In error1, \textit{lemon\_block} in the environment should be \textit{yellow\_block}; in error2, \textit{green\_block} should be placed on \textit{yellow\_block}; in error3, \textit{<move>} should be executed before \textit{<open>}).}\label{fig:error}
    \vspace{-2em}  
\end{center}
\end{figure}

Large Language Models (LLMs) have demonstrated unprecedented progress in agent planning tasks, which can generally be categorized into training paradigms~\cite{c7,c8,c9} and prompt-based paradigms~\cite{c10,c11,c12}. The former involves end-to-end training or fine-tuning of LLM parameters to directly map user instructions to robotic actions, which require costly data collection and training. As a result, an increasing number of studies favor In-Context Learning (ICL) to guide LLMs in generating appropriate output. Policy code, as one output form of ICL~\cite{c13}, refers to executable scripts composed of pre-defined API calls that directly control robotic behavior. By leveraging LLMs’ code generation capabilities, it enables mapping from natural language instructions to robot-executable actions without additional training. However, due to the inherent limitations of LLM capabilities and their context window, it remains difficult for LLMs to accurately align with key information through policy code, especially in long-horizon implicative tasks~\cite{guo2024castl}.

Specifically, when executing long-horizon implicative tasks, as shown in Fig.~\ref{fig:error}, LLMs struggle to de-implicit(i.e., resolve underspecified or ambiguous expressions into concrete, executable referents) and decompose long-horizon tasks. This often leads to the insertion of incorrect API parameters (error1) and comments (error2) when generating policy code. Additionally, when sequencing actions in long-horizon tasks, LLMs may easily confuse the correct API sequence (error3). These errors ultimately result in task failure during robot operation. Thus, we pose the following question: \textbf{how can we improve the success rate of long-horizon implicative tasks without modifying the internal parameters of the LLM?}

Inspired by how humans solve implicative problems: we first simplify complex problems into known issues before attempting to solve them, and then either summarize or discard the original approach based on the correctness of the solution. This method is also applicable to LLMs. In this paper, we propose the Triple-S framework, where the collaboration between Simplification LLM, Solution LLM, and Summary LLM significantly reduces three types of errors associated with policy code in long-horizon implicative tasks, improving the overall success rate. Specifically, the Simplification LLM simplifies instructions, including de-implicit and task decomposition. The former ensures the correct filling of API parameters, while the latter reduces errors in comments generated by LLMs. In the Solution LLM, we input the simplified instructions and retrieved top-k relevant demonstration based on similarity matching, allowing the LLM to predict policy code through ICL. In the Summary LLM, successfully predicted results are further encapsulated, including API and demonstration encapsulation. This high-level encapsulation helps reduce failures caused by API sequencing errors.

To validate the effectiveness of our framework, we introduced a new dataset called the Long-horizon Desktop Implicative Placement (LDIP) dataset, which features long-horizon tasks with implicative instructions in both observable and partially observable scenarios. Through extensive experiments, we demonstrated the effectiveness and robustness of the Triple-S framework. The main contributions of our work are as follows:
\begin{itemize}
\item We constructed the Long-horizon Desktop Implicative Placement (LDIP) dataset, featuring inference tasks on relative positioning, color, geometry, and partial observability. It serves as an effective benchmark for assessing LLMs' ability to follow long-horizon implicative instructions through policy code.
\item We proposed the Triple-S collaborative multi-LLM framework, which follows a closed-loop Simplification-Solution-Summary process to significantly reduce robot motion failures in long-horizon implicative tasks.
\item A novel demonstration library update mechanism is proposed, which encapsulates successful tasks and removes redundant demonstrations during the update process, effectively generalizing to previously failed tasks.

\end{itemize}

\section{RELATED WORK}

\subsection{In-Context Learning for Robot Planning}

Natural language-driven robot planning is a key step toward embodied intelligence. Traditional methods rely heavily on semantic parsing~\cite{c15,c16}, limiting scalability. In-Context Learning (ICL) offers a training-free paradigm~\cite{c14} that enables LLMs to generalize by leveraging semantic similarity~\cite{c11,c17} and environmental feedback~\cite{c19,c20}. While effective for mapping instructions to existing skills, these methods assume that appropriate skills already exist in the library, making them less effective for implicative instructions~\cite{c13}. In other word, clear instructions and an expandable skill library are essential for addressing these challenges. Our proposed Triple-S framework mitigates these limitations through task simplification and summarization.

\subsection{Controlling robots via policy code}
Recent work has explored using LLMs for long-horizon planning by integrating them with classical planners or task-and-motion planning systems, e.g., LLM+P~\cite{liu2023llmp}, DELTA~\cite{liu2024delta}, and AutoTAMP~\cite{chen2024autotamp}. These approaches translate natural language into structured planning representations such as PDDL and rely on external solvers for plan generation. While effective, they often require domain formalization and lack flexibility in novel environments or task variations. In contrast, the strong code-generation capabilities of LLMs~\cite{c21} have enabled direct control of robots via policy code, which bypasses symbolic planners and generates executable scripts end-to-end. For instance, ChatGPT for Robotics~\cite{c22} employs few-shot prompting to generate Python code for downstream tasks. VOYAGER~\cite{c23} uses dynamic code libraries to support autonomous exploration and skill acquisition in Minecraft. FLTRNN~\cite{c24} improves planning faithfulness through task decomposition and memory mechanisms. Code as Policies~\cite{c13} enables recursive API generation through hierarchical scaffolding, and Promptbook~\cite{c25} integrates examples, documentation, CoT reasoning, and state prediction for robust code-based planning.

While previous works have achieved notable success, they primarily target relatively simple tasks. For long-horizon implicative tasks, effective solutions via policy code remain scarce. Our framework leverages structured prompt templates to guide LLMs in generating complete policy code, enabling precise API sequencing and parameter control. By integrating ICL with collaborative LLMs, Triple-S provides a robust and generalizable solution for these challenging tasks.

\begin{figure*}[htp]
\begin{center}
\includegraphics[scale=0.535]{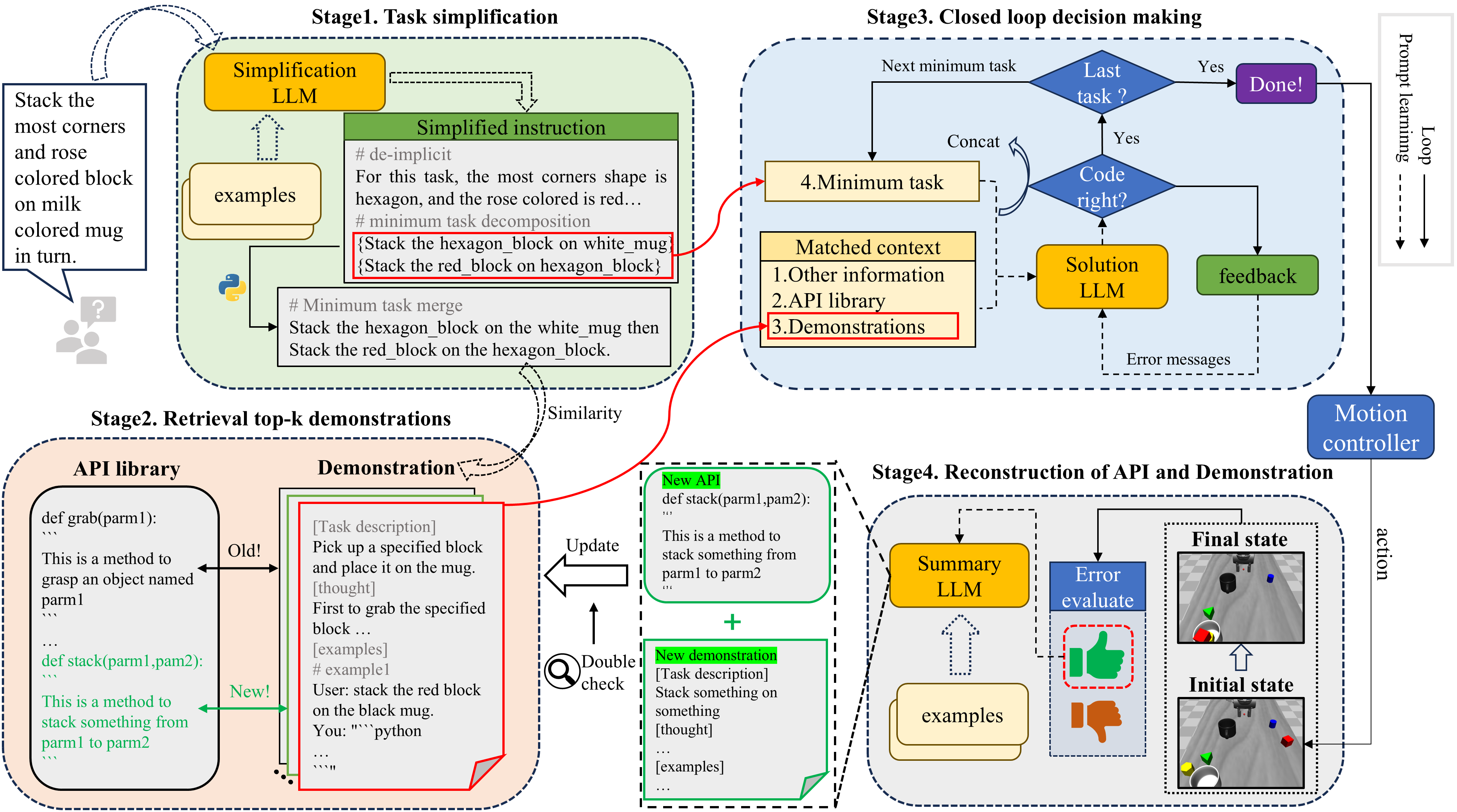}
\caption{The Triple-S Framework: it takes implicative user instructions as input and outputs executable policy code. In Stage 1, the user's long-horizon implicative instructions are simplified into de-implicit minimal task. In Stage 2, these minimal tasks retrieve the top-k relevant demonstrations. In Stage 3, the minimal tasks and demonstrations are combined for the LLM to generate policy code. In Stage 4, successfully executed tasks are further encapsulated. Detailed prompt structures are available in our code.}\label{fig:kuangjia}
    \vspace{-2em}  
\end{center}
\end{figure*}

\section{PROBLEM STATEMENT}
To better understand the execution of long-horizon tasks in the robotics domain, we define this type of problem as a tuple $\left \langle O,P,A,S,T,I,G,t \right \rangle$, where $O$ represents the set of all available objects in the environment, and $P$ represents the inherent properties of these objects. The action space $A$ includes a set of discrete high-level actions $a\in A$, which are encapsulated as high-level APIs. $S$ denotes the state space, and the transition function $T:S \times A\to S$ determines the next state of the environment based on the current state and selected action. $I$ and $G$ represent the initial state and goal state, respectively, and the objective of the task is to find an action sequence $\Pi =\left ( a_{1},a_{2},\dots,a_{n}     \right ) $ that completes the state transition. In long-horizon tasks, the number of actions $n$ is typically large~\cite{c25}. Without access to the explicit goal state $G$, the LLM relies on the task description $t$ to generate actions that drive the environment toward the intended outcome.

Long-horizon implicative tasks introduce an added layer of complexity beyond standard long-horizon tasks. Given that LLM inputs include both user instructions and environmental observations, we categorize implications into two types based on their dominant source: instruction implication and environment implication. Instruction implication arises when objects mentioned in the instruction do not clearly correspond to those in the environment. For example,if $t=``pick\ up\ the\ no-corners\ block"$ and $O = [ ``block1",``block2"], P = [ ``circle",``triangle"]$, the LLM must perform additional reasoning to generate the correct action $a = robot.pick(``block1") \ if \ block1.shape == ``circle"$. Environment implication occurs under partial observability~\cite{c26}, requiring additional information to clarify the current state, often through sensor queries. For instance, given $t = ``pick\ up\ the\ lightest \ block"$ and $O=[ ``block1",``block2"]$, the LLM must invoke a sensor API such as $get\_obj\_mass$ to obtain $P = [block1\_weight, block2\_weight]$, then execute $a = robot.pick(``block1") \ if \ block1\_weight < block2\_weight$. These implications increase the LLM's reasoning burden, especially as task volume grows. Our proposed framework addresses both implicative tasks, focusing on the LLMs' ability to solve specified planning tasks, with emphasis on task feasibility and success rate rather than optimization.


\section{METHODOLOGY}
In this section, we introduce the Triple-S, a novel framework designed to improve the success rate of long-horizon implicative tasks through multi-LLM collaboration. As shown in Fig~\ref{fig:kuangjia}, it consists of four Stages: Task Simplification, Retrieval top-k demonstrations, Closed loop decision making, and Reconstruction of API and demonstration.
\subsection{Task simplification} 
Directly inputting a long-horizon implicative task into the Solution LLM requires it to perform both instruction reasoning and code generation, which remains challenging even for advanced models. As shown in Fig.\ref{fig:error}, this often leads to errors in API parameters (error1) and comments (error2). To address this, we introduce the Simplification LLM (Stage 1, Fig.\ref{fig:kuangjia}), which transforms the original task $x_{high}$ into a de-implicit minimal task $x_{low}$ via ICL:

\begin{equation}\label{e1}
x_{low} = f_{LLM}(prompt, \ x_{high} ) 
\end{equation}
Where $prompt = <I_{sim},R_{sim},E_{sim}   >$:
\begin{itemize}
\item $I_{sim}$ represents the simplification role played by the LLM and the environmental state information.
\item $R_{sim}$ outlines the rules for simplifying the task, including: 1) identifying implication or unclear parts of the instruction and replacing them through reasoning, 2) converting the long-horizon task into a minimal task, such as \textit{"Move the gripper <length> <orientation>,"} where \textit{<length>} and \textit{<orientation>} are inferred based on the original instruction, and 3) if certain parts of the instruction cannot be replaced in steps 1 and 2, they are kept as is.
\item $E_{sim}$ provides examples of user inputs and their corresponding simplified outputs.
\end{itemize}

Through this process, the Simplification LLM minimizes the complexity of the instruction while retaining any parts that cannot be replaced. While prior works such as CaStL~\cite{guo2024castl} have explored instruction de-implicitation via symbolic translation, our approach performs it in a code generation context and further integrates it with task decomposition for long-horizon execution.

\subsection{Retrieval top-k demonstration}
When all task demonstrations are provided as context to the LLM, it often fails to capture key demonstration for specific tasks~\cite{c14}, leading to execution failures. This issue worsens as the demonstration library expands.

Given the successful application of Retrieval Augmented Generation (RAG) in natural language processing, we adopt it to retrieval the top-k demonstrations to filter out irrelevant ones. Specifically, we combine the minimal tasks from Stage1 and use the \textit{all\_datasets\_v4\_MiniLM-L6} embedding model to embed it's content as query $q$. Each demonstration's \textit{[task description]} is embedded as a key $d$, and by calculating the cosine similarity between the query and each key, $sim(q,d_{i} )$, we retrieve the top-k relevant demonstrations:
\begin{equation}\label{e2}
D_{k} =Top\_k (sim(q,\textbf{D})
\end{equation}
Where $\textbf{D}=\left \{ d_{1},d_{2},\cdots ,d_{n}    \right \}$ represents the set of retrieved demonstrations, with $n$ being the total number. Each demonstration includes a \textit{[task description]}, \textit{[thought]}, and \textit{[examples]}, as shown on the right side of Stage2 in Fig.~\ref{fig:kuangjia}.

\subsection{Closed loop decision making}
Building on the Stage1 and Stage2, the Solution LLM receives a context with a de-implicit minimal task $x_{low}$ and the relevant demonstrations $D_{k}$. To further enhance the confidence in the generated code, we follow the prompt structure from Promptbook~\cite{c25} by embedding other information $c$ and the API library as text in the prompt. The information $c$ includes the following:
\begin{itemize}
\item The current state of the robot and its constraints.
\item Information about objects in the environment and their attributes.
\end{itemize}
The Solution LLM will then, through ICL, generate the policy code based on the given task $x_{low}$:
\begin{equation}\label{e3}
Code = f_{LLM}([c,API,D_{k},x_{low}  ])
\end{equation}
where $[\cdot  ,\cdot ]$ denotes concatenation in the specified order.

The generated code will be executed by a compiler, and if there are any syntax errors or invalid parameters, the error messages will be feedback to the Solution LLM, prompting a new round of code generation. Once all minimal tasks have been processed, the complete code will be passed to the motion controller to execute the corresponding actions.
\subsection{Reconstruction of API and Demonstration}\label{4.4}
Although the above process mitigates API parameter and comment errors, it does not fully prevent execution failures caused by incorrect API sequencing. Such errors can lead to exceeding workspace limits, collisions, or even safety hazards. To address this, we introduce the Summary LLM, which encapsulates successful executions into higher-level APIs and dynamically updates the demonstration library.

For $Code$ without errors, we need to verify whether it successfully executes the task, which is determined based on error evaluation. Suppose the environment contains $m$ objects $\left \{ o_{1},o_{2},\cdots ,o_{m}    \right \} $, with each object's final state vector represented as $S_{(o_{i} )} $ and its target state vector as $G_{(o_{i} )} $. The state error of object $o_{(i)} $ is denoted as $E_{(o_{i} )}$, and the gripper's state error as $E_{grip} $. The task is considered successful if it meets the following condition:
\begin{equation}\label{e4}
\sum_{i=1}^{m}E_{(o_{i} )} +E_{grip}\le \varepsilon  
\end{equation}
Where, $E_{(o_{i} )} =\left \| S_{(o_{i} )} -G_{(o_{i} )}  \right \| _{2}$, $\varepsilon$ is the error threshold. $Code$ that satisfies Eq.~\ref{e4} will be passed, along with the de-implicit minimal task $x_{low}$, to Summary LLM for generating newly encapsulated API and demonstration:
\begin{equation}\label{e5}
API,Demonstration = f_{LLM}(prompt,x_{low},Code) 
\end{equation}
Where $prompt = <I_{sum},R_{sum},E_{sum}   >$:

\begin{figure*}[htp]
\begin{center}
\includegraphics[scale=0.53]{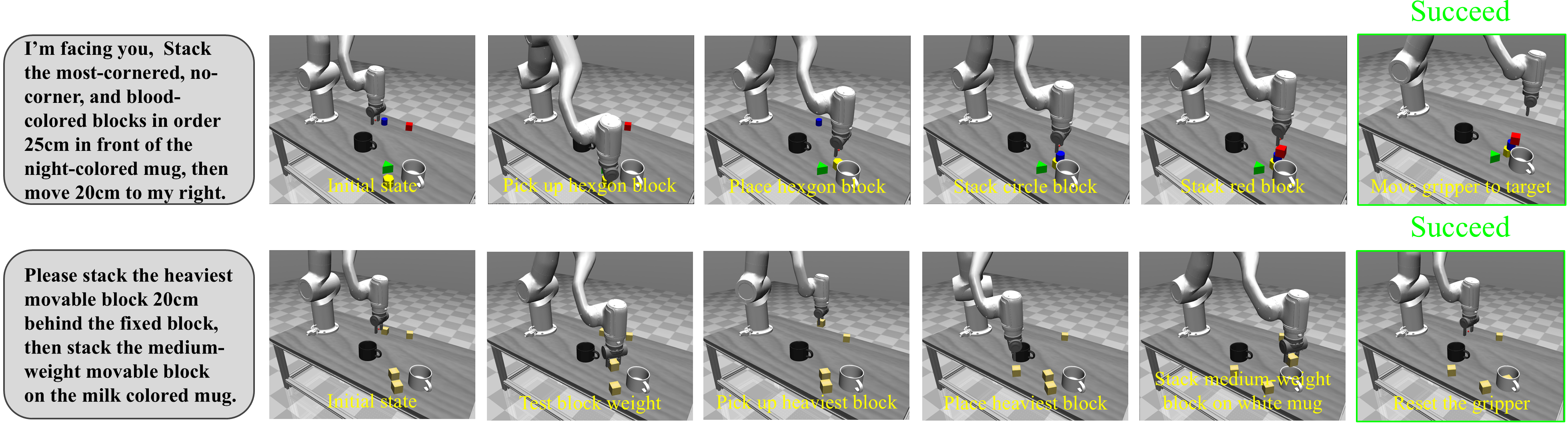}
\caption{ Key actions of the Triple-S framework while executing long-horizon implicative tasks (only critical actions are displayed).}\label{fig:move}
    \vspace{-1em}  
\end{center}
\end{figure*}

\begin{table*}[ht]
    \renewcommand{\arraystretch}{1.25}
    \caption{Performance of different policy code methods on the LDIP, Our method achieved the best results across all metrics.} 
    \label{tab:t1}    

\centering
\tabcolsep=0.025\linewidth    
\begin{tabular}{llcccccc}
\hline
\multicolumn{1}{c}{\multirow{2}{*}{Models}} & \multicolumn{1}{c}{\multirow{2}{*}{Methods}} & \multicolumn{3}{c}{Observable}                              & \multicolumn{3}{c}{Partially observable}                     \\ \cline{3-8} 
\multicolumn{1}{c}{}                        & \multicolumn{1}{c}{}                         & SR↑                 & Err↓               & ESR↑                & SR↑                 & Err↓               & ESR↑                \\ \hline
\multirow{4}{*}{Llama3}                     & CFR                                          & 48.88±0.67          & 0.319±0.002          & 53.81±0.34          & 43.30±2.06          & 0.309±0.009          & 45.88±1.43          \\
                                            & FLTRNN                                       & 60.27±0.45          & 0.273±0.011          & 67.67±0.33          & 51.04±0.05          & 0.306±0.007          & 51.57±1.06          \\
                                            & VOYAGER                                      & 57.37±1.12          & 0.366±0.007          & 69.08±0.23          & 68.56±0.05          & 0.165±0.024          & 76.46±1.18          \\
                                            & Ours                                         & \textbf{84.38±0.89} & \textbf{0.069±0.003} & \textbf{86.52±1.91} & \textbf{81.96±0.05} & \textbf{0.134±0.005} & \textbf{91.38±0.57} \\ \hline
\multirow{4}{*}{GPT3.5}                     & CFR                                          & 82.37±2.46          & 0.083±0.023          & 82.37±2.46          & 73.71±2.58          & 0.068±0.058          & 74.08±2.21          \\
                                            & FLTRNN                                       & 87.05±0.45          & 0.064±0.003          & 87.25±0.64          & 89.18±0.05          & 0.032±0.001          & 90.58±0.05          \\
                                            & VOYAGER                                      & 83.48±3.57          & 0.085±0.028          & 84.21±2.84          & 89.18±0.05          & 0.050±0.011          & 89.64±0.98          \\
                                            & Ours                                         & \textbf{93.75±0.41} & \textbf{0.027±0.002} & \textbf{93.96±0.24} & \textbf{95.88±1.03} & \textbf{0.029±0.013} & \textbf{96.87±0.03} \\ \hline
\end{tabular}
    \vspace{-1.5em}  
\end{table*}

\begin{figure}[htp]
\begin{center}
\includegraphics[scale=0.47]{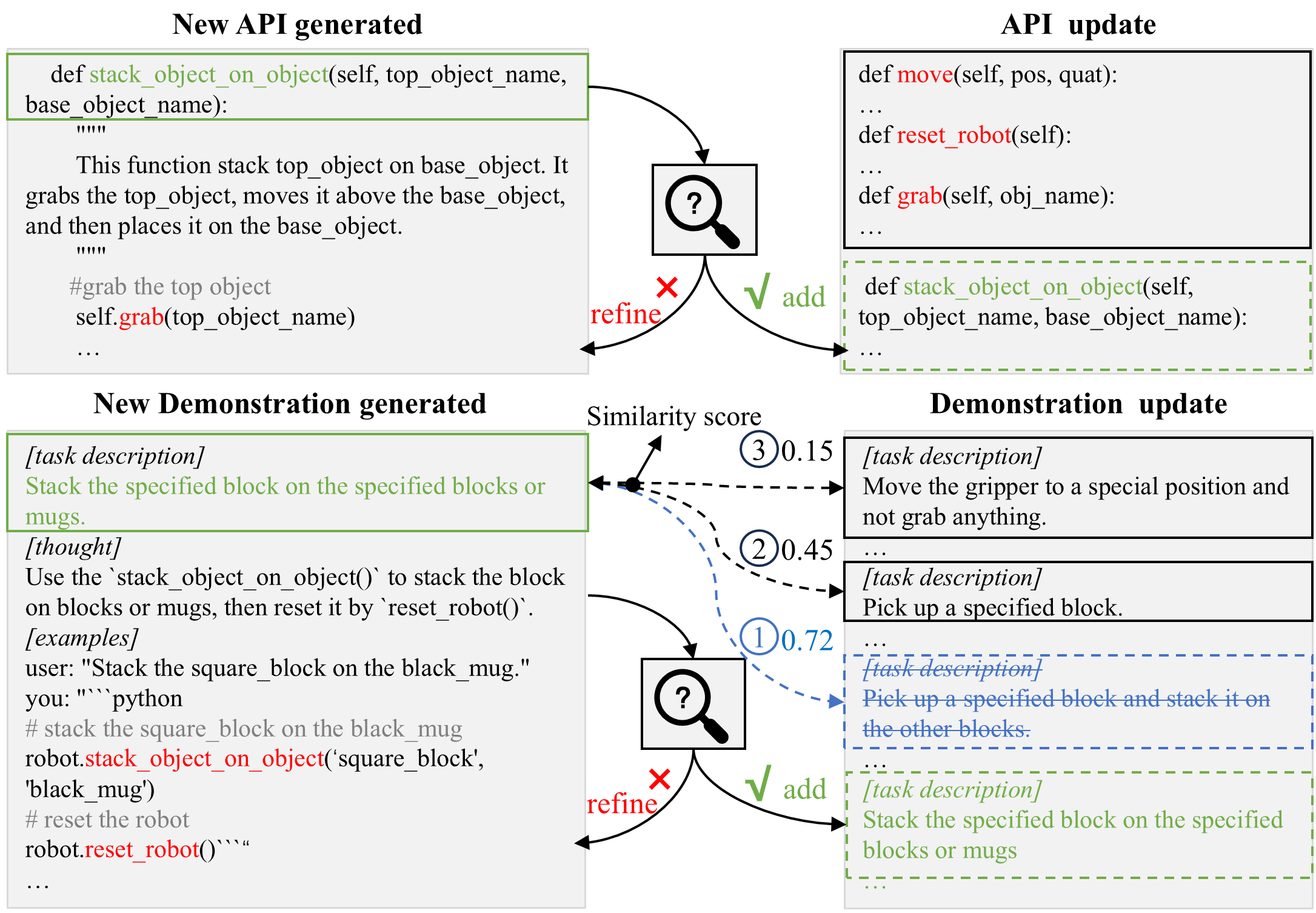}
\caption{The update of API and Demonstration}\label{fig:skill}
    \vspace{-3em}  
\end{center}

\end{figure}
\begin{figure}[htp]
\begin{center}
\includegraphics[scale=0.5]{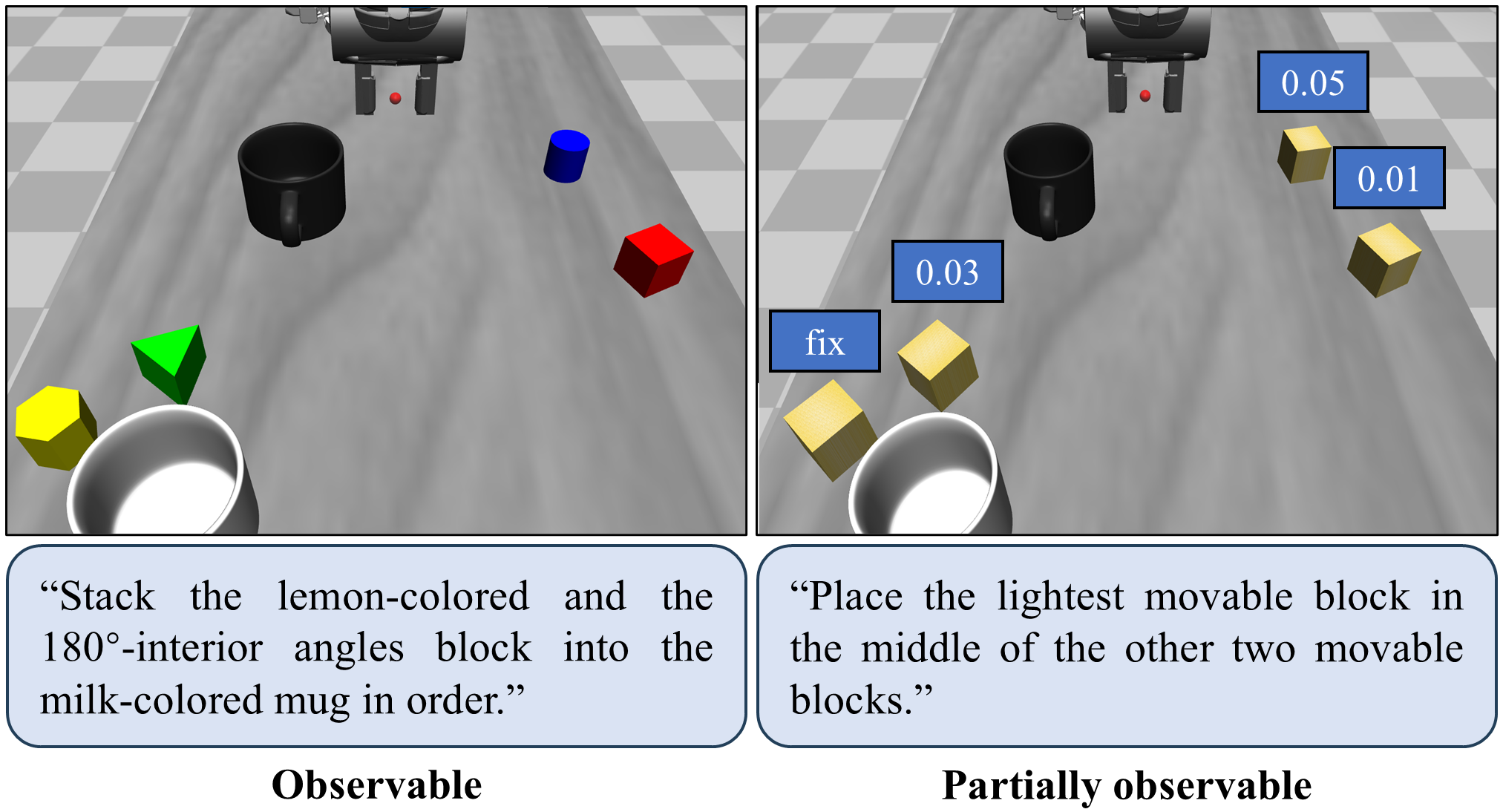}
\caption{Examples of environment observations and linguistic instructions}\label{fig:dataset}
    \vspace{-2.5em}  
\end{center}
\end{figure}
\begin{itemize}
\item $I_{sum}$ represents the summarization role played by the LLM and the environmental state information.
\item $R_{sum}$ outlines the rules for summarization, including: 1)The purpose of encapsulation is to simplify the original code while retaining task functionality, 2)Further encapsulation must be built on the existing APIs and 3)Example encapsulation must include the creation of \textit{[task description]}, \textit{[thought]}, and \textit{[examples]}.
\item $E_{sum}$ provides examples of user inputs and their corresponding encapsulation.
\end{itemize}

We utilize the Summary LLM to encapsulate successfully executed code, and the encapsulated API and demonstration are further verified by an additional Summary LLM acting as a supervisor, which has been proved as effective in VOYAGER~\cite{c23}. Unlike their approach, our method involves not only adding the newly encapsulated APIs but also dynamically updating the demonstration library. As shown in Fig.~\ref{fig:skill}, new demonstration undergo similarity-based retrieval against the original demonstration library. If the \textit{[task description]} similarity exceeds a given threshold, the new demonstration is considered to correspond to the same task category, and the original demonstration is removed during the update process. This dynamic updating prevents Stage2 from retrieving outdated demonstration that lack the updated API in their \textit{[examples]} section. Moreover, we hypothesize that newly encapsulated APIs and updated demonstrations from successful tasks can assist in solving previously failed ones. While such encapsulations simplify code generation, the underlying task complexity remains: these high-level APIs are not predefined, but progressively learned from low-level executions. As shown in Fig.~\ref{fig:api}, some long-horizon implicative tasks remain unsolved without encapsulation.

\section{EXPERIMENTS}

In this section, we first introduce the experimental setup, including the Long-horizon Desktop Implicative Placement (LDIP) dataset, and then evaluate the effectiveness of the Triple-S framework on this dataset.
\subsection{Experiments setup}
\textbf{Datasets.} To evaluate the Triple-S framework, we introduce LDIP, a new dataset built on the Robopal simulation environment~\cite{c28}, as no existing dataset explicitly targets long-horizon implicative tasks. As illustrated in Fig.~\ref{fig:dataset}, LDIP includes both observable and partially observable scenarios. The observable subset comprises 500 tasks, over half involving instruction implication, requiring reasoning over relative position, color, and shape. It includes four blocks of varying colors and shapes, and two cups of different colors. Here, the LLM must simplify instructions to correctly align linguistic references with available objects. The partially observable subset includes 97 tasks involving environment implication, with three movable blocks of different weights, one fixed block, and two colored cups. Since visual models cannot distinguish the blocks by mass, gravity sensor queries (e.g., \textit{get\_obj\_mass}) are required, increasing the code generation burden. Each instruction is annotated with corresponding policy code and final object positions. We manually encapsulated a set of core APIs (\textit{Pick, Place, Move, get\_obj\_pose, get\_obj\_mass}), and verified each instruction’s correctness against its final state to ensure data quality.

\textbf{Methods.} In simulation experiments, We compared three methods that use policy code to drive agents: ChatGPT for Robotics (CFR)~\cite{c22}: A strategy combining ICL and Chain-of-Thought reasoning. FLTRNN~\cite{c24}: A method combining ICL, task decomposition, and rule injection, excelling in VirtualHome. VOYAGER~\cite{c23}: A strategy that combines ICL with dynamic skill library updates, excelling in Minecraft. All methods, including Triple-S, were evaluated using two representative LLMs: the open-source LLaMA3-8B-Instruct (Llama3) and the commercial GPT-3.5-Turbo-0613 (GPT3.5), reflecting a trade-off between model capability and token cost.

\textbf{Metrics.} Following prior benchmarks~\cite{c6,c11}, we report the standard Success Rate (\textit{SR}), which is the proportion of tasks that are successfully completed. In addition, we define two auxiliary metrics to better reflect execution quality: Error (\textit{Err}): Measures the average deviation between the final environment state and the ground truth, based on object states and gripper state (as described in Eq.~\ref{e4}). Executable Success Rate (\textit{ESR}): The proportion of successfully completed tasks among those whose generated code can be executed without syntax or API errors. Notably, \textit{SR} = 1 only if \textit{ESR} = 1.

\subsection{Results}
We conducted three tests on the LDIP dataset to obtain the average of all metrics. The main experimental results are shown in Tab.~\ref{tab:t1}, and we found the following: 1) The Triple-S framework achieved state-of-the-art results across all metrics, demonstrating its effectiveness in solving long-horizon tasks involving both instruction and environment implication. Fig.~\ref{fig:move} illustrates the key motion sequences of Triple-S framework when executing these tasks. 2) Compared to other methods, Triple-S exhibited the lowest variance across multiple metrics, indicating strong stability and robustness. 3) As model capacity increased, all methods exhibited improved performance. Notably, Triple-S still achieved strong results even with the smaller LLaMA3-8B. In both observable and partially observable scenarios, it outperformed the best existing methods, with \textit{SR} gains of 24.11\% and 13.4\%, respectively. 4) FLTRNN performed well in observable scenarios, likely due to the effectiveness of its task decomposition module. Conversely, VOYAGER excelled in partially observable scenarios, likely owing to its dynamic skill library update mechanism. However, neither method achieved a comprehensive advantage across all scenarios, underscoring the importance of combining task simplification with task summarization.
 
\section{ANALYSIS AND DISCUSSION}
We conducted a more detailed analysis of the Triple-S framework, from robustness, ablation study, update mechanism to generalization. Unless otherwise noted, the results are weighted averages from both the Llama3 and GPT3.5.

\subsection{Robustness with different complexity of task}
\begin{figure}[t]
    \centering
    \includegraphics[width=0.235\textwidth]{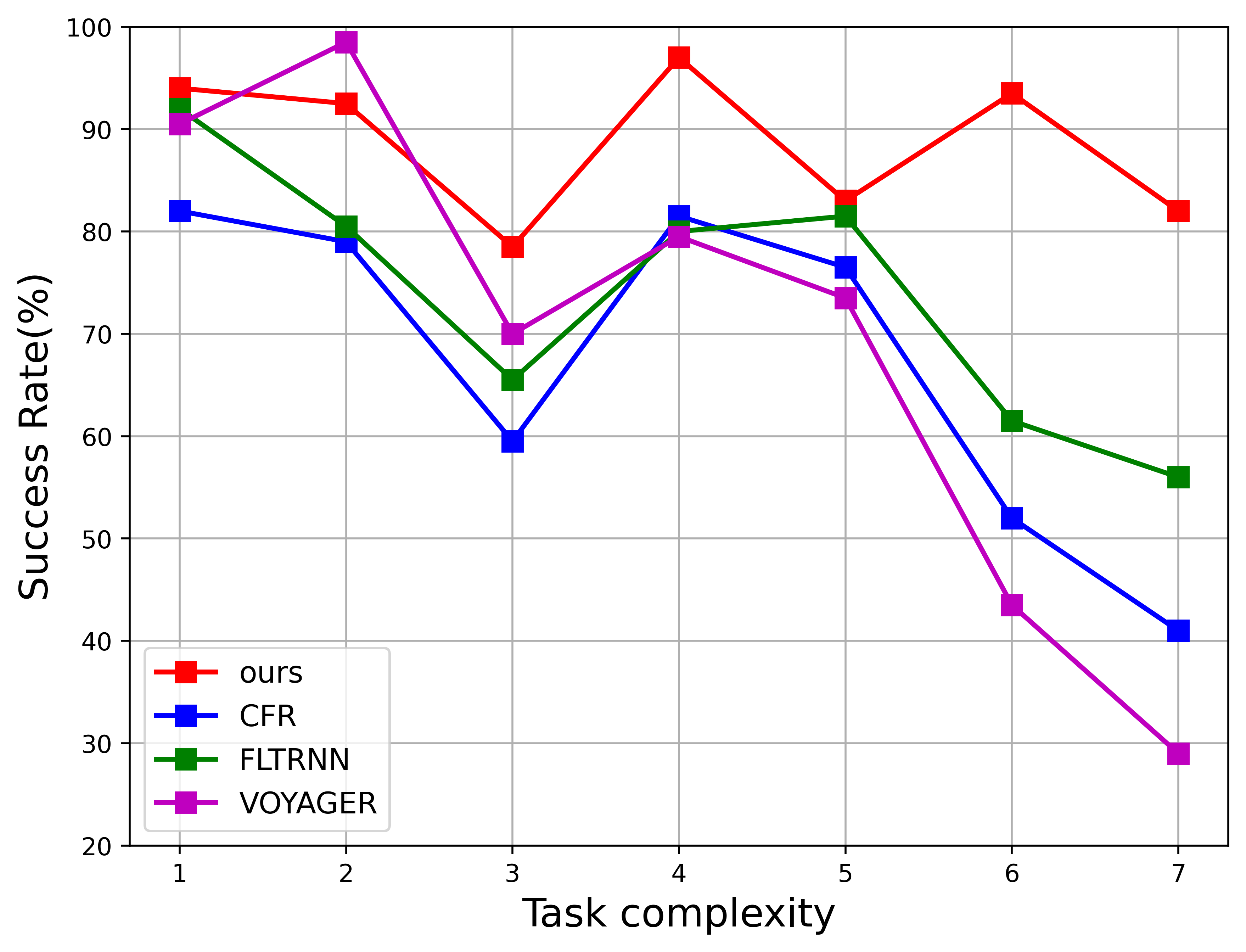}
    \includegraphics[width=0.235\textwidth]{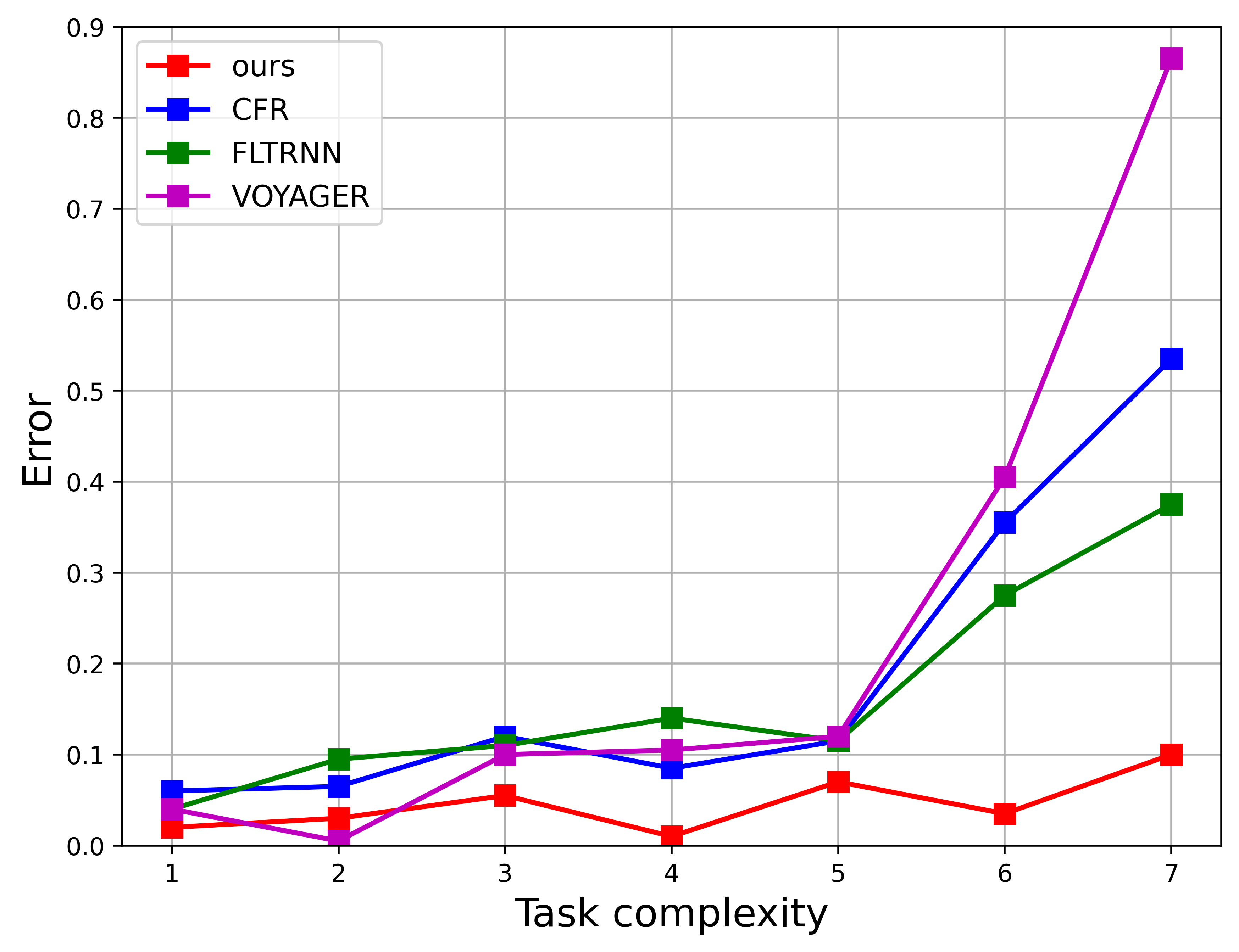}
    \caption{Performance of each method as task complexity increases}\label{complex}
        \vspace{-0em}  
\end{figure}

To evaluate robustness under different task complexities, we assigned complexity levels to tasks in the LDIP dataset based on the degree of implication and the number of required APIs. Tasks were categorized into seven levels, with higher levels denoting increased difficulty, and these levels were validated against human judgment. As shown in Fig.~\ref{complex}, the results indicate: 1) Triple-S exhibited strong robustness, with minimal variation across levels—with the maximum variation in \textit{SR} and \textit{Err} being only 18.5$\%$ and 0.09, respectively. 2) Other methods suffered from increasing performance degradation as complexity rose, particularly beyond level 5, where both \textit{SR} and \textit{Err} deteriorated significantly. These findings suggest that existing methods struggle with high-complexity tasks, while Triple-S, through multi-LLM collaboration, maintains high success (\textit{SR}=82$\%$) and low error (\textit{Err}=0.1) even on the most challenging tasks.


\begin{figure}[t]
\begin{center}
\includegraphics[scale=0.31]{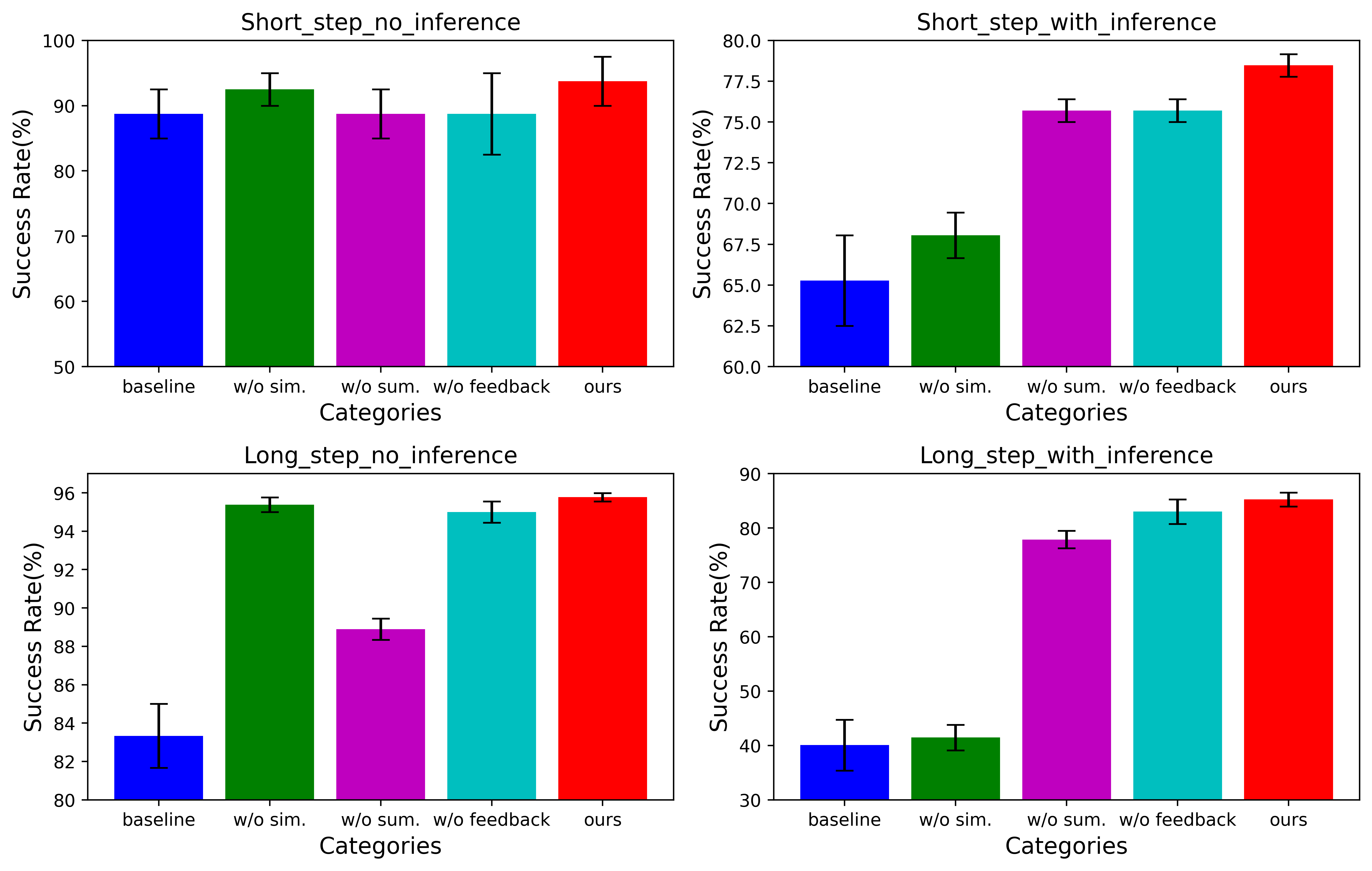}
\caption{Ablation performance of the Triple-S on the four sub-tasks.}\label{fig:ablation}
    \vspace{-2.5em}  
\end{center}
\end{figure}

\subsection{Ablation study}
To assess the contribution of each module in Triple-S, we conducted ablations on 500 observable-environment tasks, divided into four subgroups based on task length and reasoning demand, using CFR as the baseline. As shown in Fig.~\ref{fig:ablation}, and we found the following: 1) Removing the Simplification LLM led to significant drops in reasoning tasks (\textit{SR} declines of 10.42$\%$ for short-horizon, and 43.81$\%$ for long-horizon), confirming its key role in instruction decomposition and de-implication. 2) Excluding the Summary LLM caused \textit{SR} declines of 6.88$\%$ and 7.38$\%$ in long-horizon tasks, demonstrating its importance in encapsulating reusable high-level APIs. 3) The feedback module yielded modest but consistent improvements (2.7$\%$ average), aiding correction in borderline cases. These results highlight that each component contributes distinct and complementary value, especially under long-horizon and reasoning-intensive conditions.

\begin{figure*}[htbp]
\begin{center}
\includegraphics[scale=0.53]{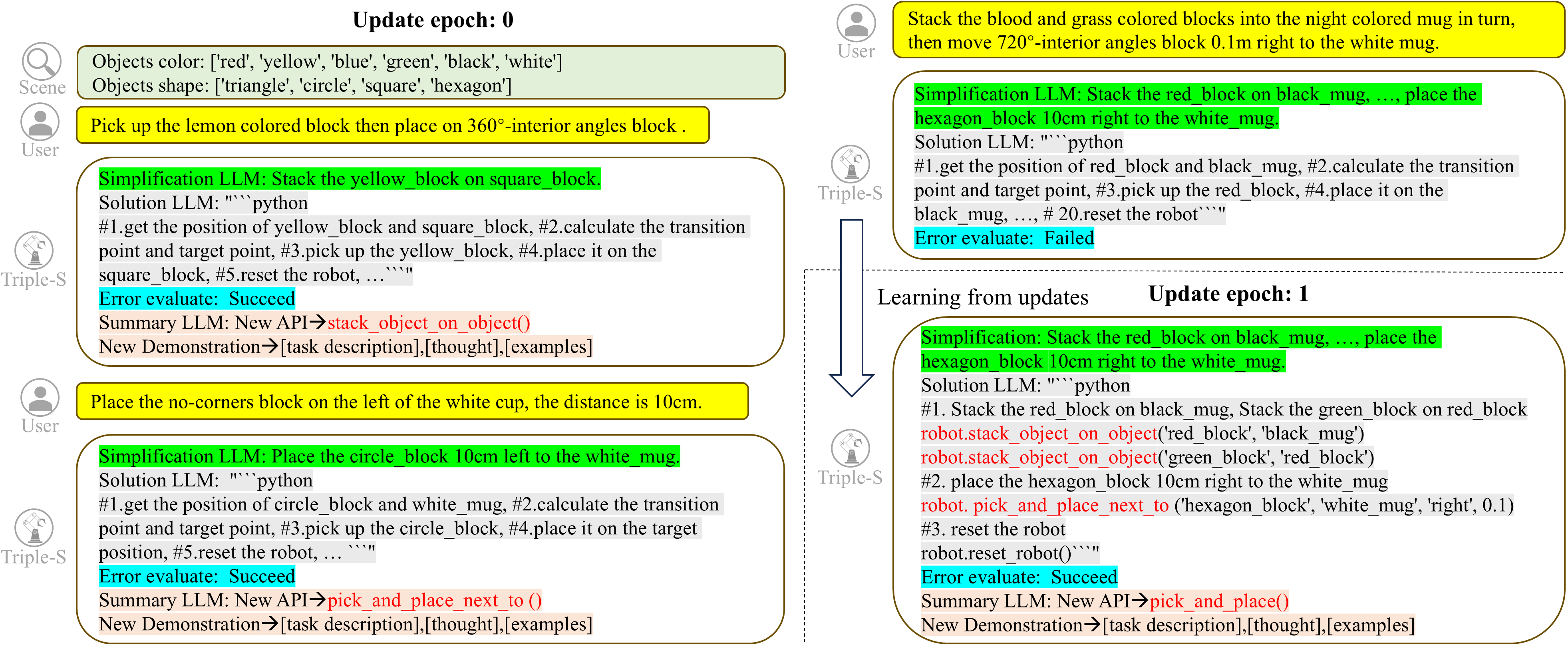}
\caption{Encapsulation from successful tasks assist previously failed tasks.}\label{fig:api}
    \vspace{-1.5em}  
\end{center}
\end{figure*}

\begin{table}[t]
\renewcommand{\arraystretch}{1.25}
\begin{threeparttable}    
    \centering
    \caption{Impact of different demonstration library update methods.} 
    \label{tab:t2}    
    \tabcolsep=0.04\linewidth 
\begin{tabular}{ccccc}
\hline
\multicolumn{3}{c}{Update configuration} & \multirow{2}{*}{SR} & \multirow{2}{*}{Err} \\ \cline{1-3}
Epochs       & Append       & Delete      &                     &                        \\ \hline
0           & ×            & ×           & 76.50±1.00          & 0.086±0.017            \\
1           & √            & ×           & 80.50±0.50          & 0.070±0.003            \\
1           & √            & √           & 89.00±0.50          & 0.053±0.003            \\
2           & √            & √           & \textbf{91.25±0.25} & \textbf{0.040±0.001}   \\ 
3           & √            & √           & 90.21±0.50          & 0.062±0.004            \\ \hline
\end{tabular}
\begin{tablenotes}[para,flushleft]  
        \item "Append" indicates the addition of new demonstration to the library, while "Delete" refers to the removal of demonstration from the library.
     \end{tablenotes} 
\end{threeparttable} 
    \vspace{-2em}  
\end{table}
\subsection{Impact of demonstration library updates}
To further evaluate the impact of the demonstration update mechanism (Section~\ref{4.4}) in the Triple-S framework, we randomly selected 100 test samples from LDIP and evaluated each sample three times, the results are shown in Tab.~\ref{tab:t2}. We found: (1) Demonstration updates are beneficial. Compared to the non-updated demonstrations, using the Append-only and Append-Delete mechanisms improve the SR of 4$\%$ (row 1, row 2) and 12.5$\%$ (row 1, row 3), respectively. (2) The Append-Delete mechanism achieved higher \textit{SR} and lower \textit{Err} compared to Append-only (row 2, row 3). This is because the Append-only mechanism can still match original examples through similarity, causing the Solution LLM to follow the outdated API order, increasing the prediction difficulty. (3) As the number of update epochs increases, \textit{SR} reaches a threshold (row 1, row 3, row 4, row 5), indicating that both under- and over-encapsulation struggle to adapt to tasks of varying complexity. Noted that although epoch updates can only be performed in simulation (with labels), the optimal results from iteration can serve as a reference for actual robot motion.

Additionally, we investigated whether the demonstration update mechanism could improve performance on previously failed tasks, which traditionally requires substantial human intervention~\cite{c27}. As shown in Fig.~\ref{fig:api}, our findings support this hypothesis. When the update epoch was 0, Triple-S successfully completed simpler tasks such as stacking and placement, leading to the creation of new high-level APIs, such as $stack\_object\_on\_object()$ and $pick\_and\_place\_next\_to()$. However, for more complex long-horizon implicative tasks (e.g., the right panel of Fig.~\ref{fig:api}), Solution LLM still struggled—despite instruction simplification—producing up to 20 redundant steps and failing the task. Notably, no new API was encapsulated for this specific task. After one update epoch, Solution LLM was able to generalize by reusing previously encapsulated APIs, completing the task with fewer comments and reduced code generation effort. This demonstrates that the Triple-S framework can effectively support generalization to previously failed tasks by learning from updates.

\begin{figure}
	\begin{minipage}[b]{0.57\columnwidth}
		\centering
		\subfigure[][Real-world robot environment]{\includegraphics[width=1\linewidth,height=4.2cm]{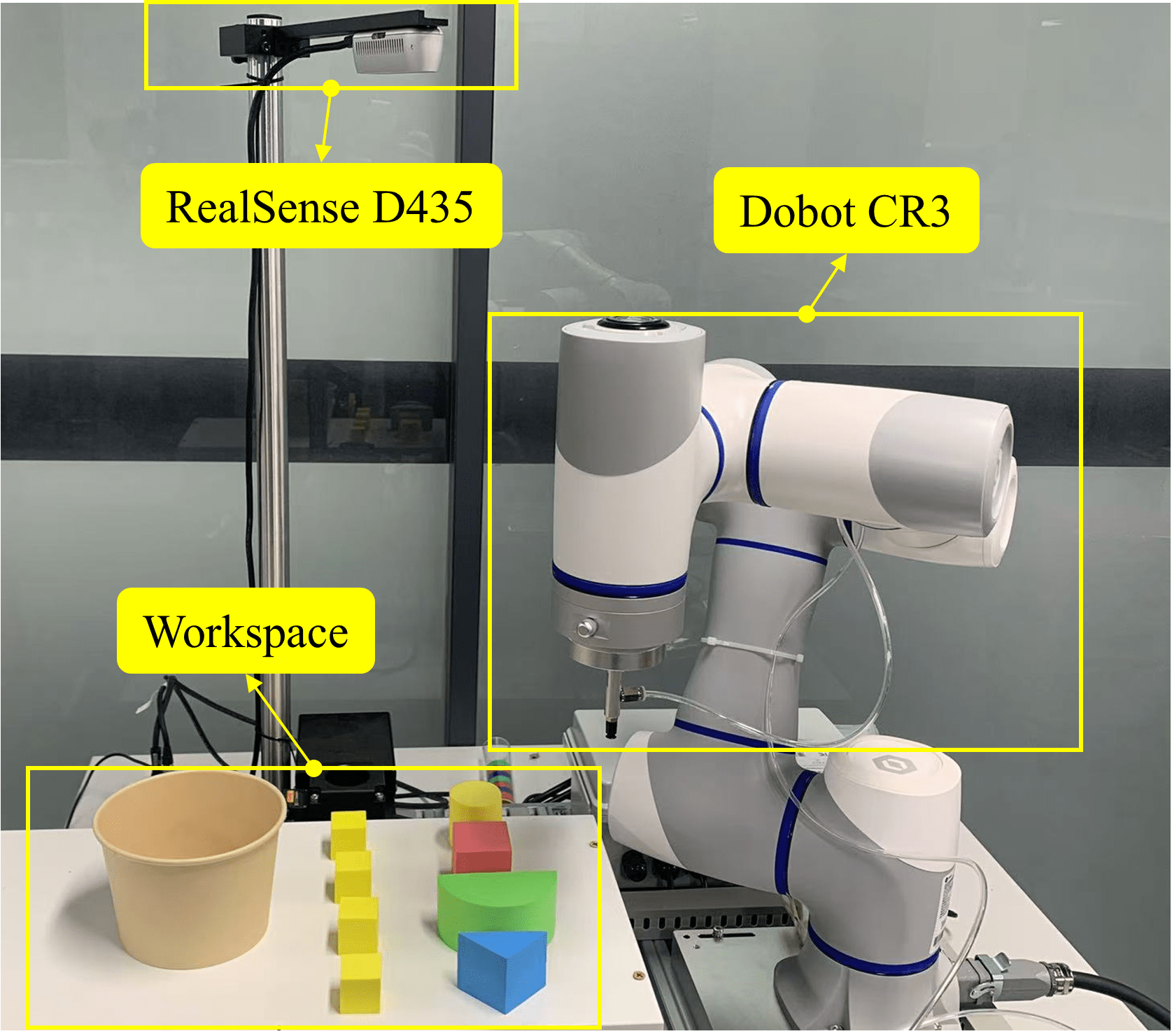}}
	\end{minipage}
	\begin{minipage}[b]{0.42\columnwidth}
		\centering
		\subfigure[][Instruction implication]{\includegraphics[width=1\linewidth,height=1.7cm]{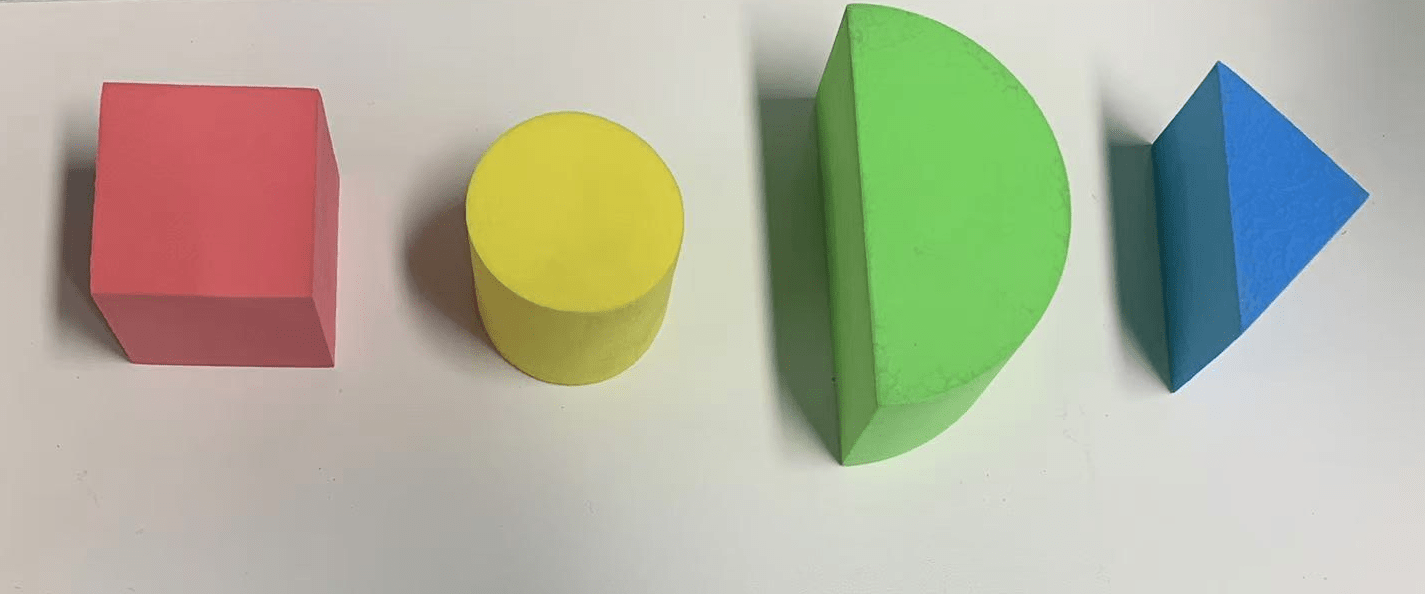}}
		\\
		\subfigure[][Environment implication]{\includegraphics[width=1\linewidth,height=1.7cm]{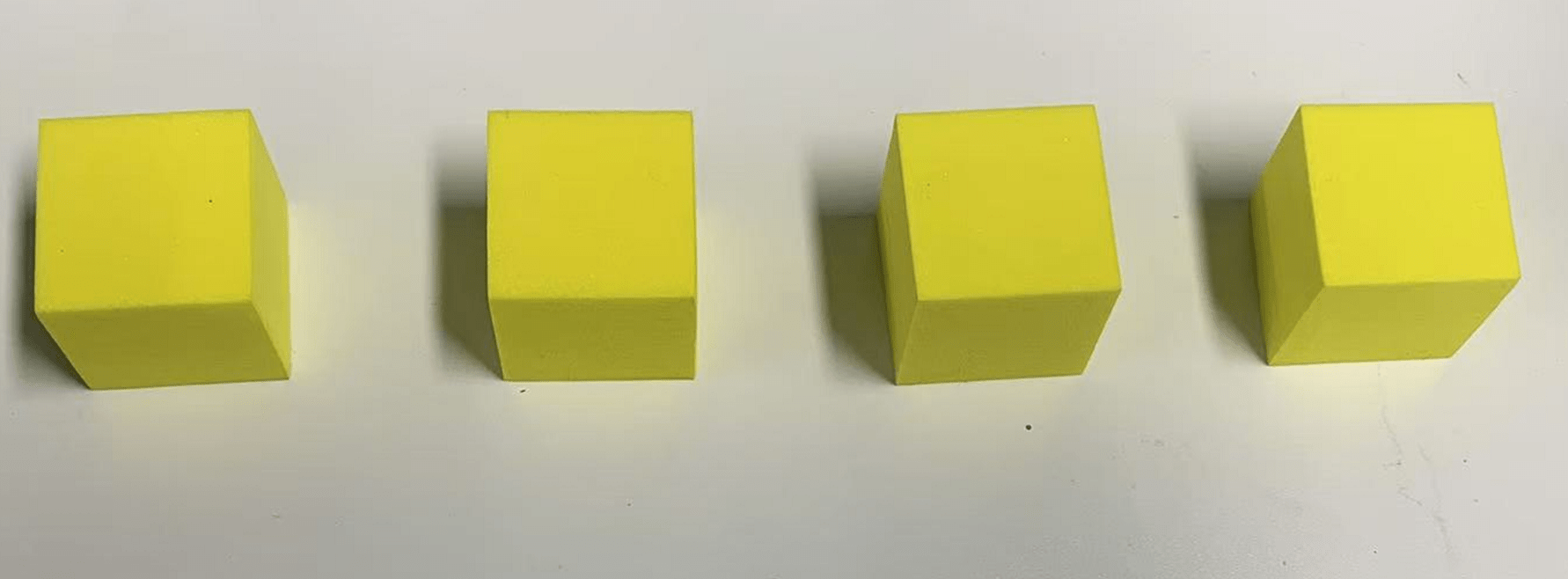}}
	\end{minipage}
	\caption{Experiment set up}\label{real robot}
    \vspace{-0.8em}  
\end{figure}

\begin{table}[t]
\renewcommand{\arraystretch}{1.25}
    \caption{Success rate of task execution on the real-world robot.} 
    \label{tab:t4}    
        \centering
            \tabcolsep=0.032\linewidth 
\begin{tabular}{ccccccc}
\hline
\multirow{2}{*}{Task} & \multicolumn{2}{c}{Code as Policy} & \multicolumn{2}{c}{Promptbook} & \multicolumn{2}{c}{Triple-S}   \\ \cline{2-7} 
                      & Ins.          & Env.         & Ins.          & Env.          & Ins.            & Env.           \\ \hline
Object-stack          & 4/10         & 5/10        & 5/10         & 5/10         & \textbf{8/10}  & \textbf{7/10} \\
Object-place          & 6/10         & 6/10        & 6/10         & 7/10         & \textbf{10/10} & \textbf{8/10} \\
Integration           & 2/10         & 3/10        & 3/10         & 4/10         & \textbf{7/10}  & \textbf{5/10} \\ \hline
\end{tabular}
    \vspace{-1.5em}  
\end{table}

\subsection{Generalization of real-world robot}
We deployed the Triple-S framework on a Dobot CR3 6-axis robotic arm to evaluate its generalization capability. An RGB-D sensor captured the raw depth map, which was then processed to generate height maps and object attributes, including color and shape. This information was integrated into the LLMs' context along with basic API prompts. We constructed two types of implication tasks consistent with the simulation experiments—Instruction implication (Ins.) and Environment implication (Env.). Each implication task is further divided into three subtasks—object stacking, object placement, and integration (a combination of the first two). Fig.~\ref{real robot} illustrates the experimental setup.

In the experiments, we selected Code as Policies~\cite{c13} and Promptbook~\cite{c25} as real-robot baselines, as they are explicitly designed for code-based control in physical environments. All frameworks were evaluated using GPT3.5 with 10 trials per task. And the depth camera was triggered only once per task, as the robotic arm can infer the placed block's position from the current gripper state, ensuring operational continuity. The results are shown in Tab.~\ref{tab:t4}. Triple-S achieved the highest performance across all tasks, which using the same demonstrations and APIs as in the simulation after 1 epoch updates, highlighting strong portability from simulation to reality. However, compared to simulation, success rates were slightly lower in real-world experiments, particularly for the object-stacking task. This was partly due to environmental perception errors, causing slight deviations in target coordinates, which occasionally exceeded precision limits for accurate manipulation, especially with smaller objects, which will improve with advancements in visual models~\cite{c29}. And we also measured end-to-end latency using locally deployed models. The average response time was 2.33 seconds across the three stages—0.88s for task simplification, 0.13s for demonstration retrieval, and 1.32s for policy code generation—demonstrating the practicality of Triple-S for real-time robotic control.

\section{CONCLUSIONS}

This paper proposes a policy code generation framework based on Large Language Models (LLMs) to address challenges with API parameters, comments, and sequencing errors in long-horizon implicative tasks. We introduce Triple-S (Simplification–Solution–Summary), a training-free, multi-LLM framework that leverages in-context learning to generate robust policy code. Its effectiveness was validated in both simulation and real-world settings using the LDIP dataset, covering diverse instruction and environment implications. While Triple-S demonstrates strong performance on tabletop block tasks, we have not yet evaluated its generalization to other domains due to reliance on task-specific environments and APIs. Nonetheless, the modular LLM roles and structured prompt format make it theoretically extensible to broader tasks via API and prompt adaptation, which we will explore in future work.












\end{CJK}

\end{document}